\pdfoutput=1

\documentclass[11pt]{article}

\usepackage{ACL2023}

\usepackage{times}
\usepackage{float}
\usepackage{latexsym}
\usepackage{graphicx} 
\usepackage{subfigure}
\usepackage{longtable}

\usepackage{stfloats}
\usepackage{arydshln}
\usepackage{booktabs}
\usepackage{multirow}
\usepackage[T1]{fontenc}

\usepackage[utf8]{inputenc}

\usepackage{microtype}

\usepackage{inconsolata}

\title{Exploiting Pseudo Image Captions for Multimodal Summarization}

\author{{Chaoya Jiang$^{1}$, Rui Xie$^{1}$, Wei Ye$^1$\thanks{ ~corresponding author.}, Jinan Sun$^{1}$,  Shikun Zhang$^{1}$} \\
$^1$ National Engineering Research Center for Software Engineering, Peking University \\
\{jiangchaoya, wye\}@pku.edu.cn,\\}

\begin{document}
\maketitle
\begin{abstract}
Multimodal summarization with multimodal output (MSMO) faces a challenging semantic gap between visual and textual modalities due to the lack of reference images for training. Our pilot investigation indicates that image captions, which naturally connect texts and images, can significantly benefit MSMO. However, exposure of image captions during training is inconsistent with MSMO's task settings, where prior cross-modal alignment information is excluded to guarantee the generalization of cross-modal semantic modeling. To this end, we propose a novel coarse-to-fine image-text alignment mechanism to identify the most relevant sentence of each image in a document, resembling the role of image captions in capturing visual knowledge and bridging the cross-modal semantic gap. Equipped with this alignment mechanism, our method easily yet impressively sets up state-of-the-art performances on all intermodality and intramodality metrics (e.g., more than 10\% relative improvement on image recommendation precision). Further experiments reveal the correlation between image captions and text summaries, and prove that the pseudo image captions we generated are even better than the original ones in terms of promoting multimodal summarization.
\end{abstract}

\section{Introduction}

\begin{figure}[h]
\centering
\includegraphics[width=0.45\textwidth]{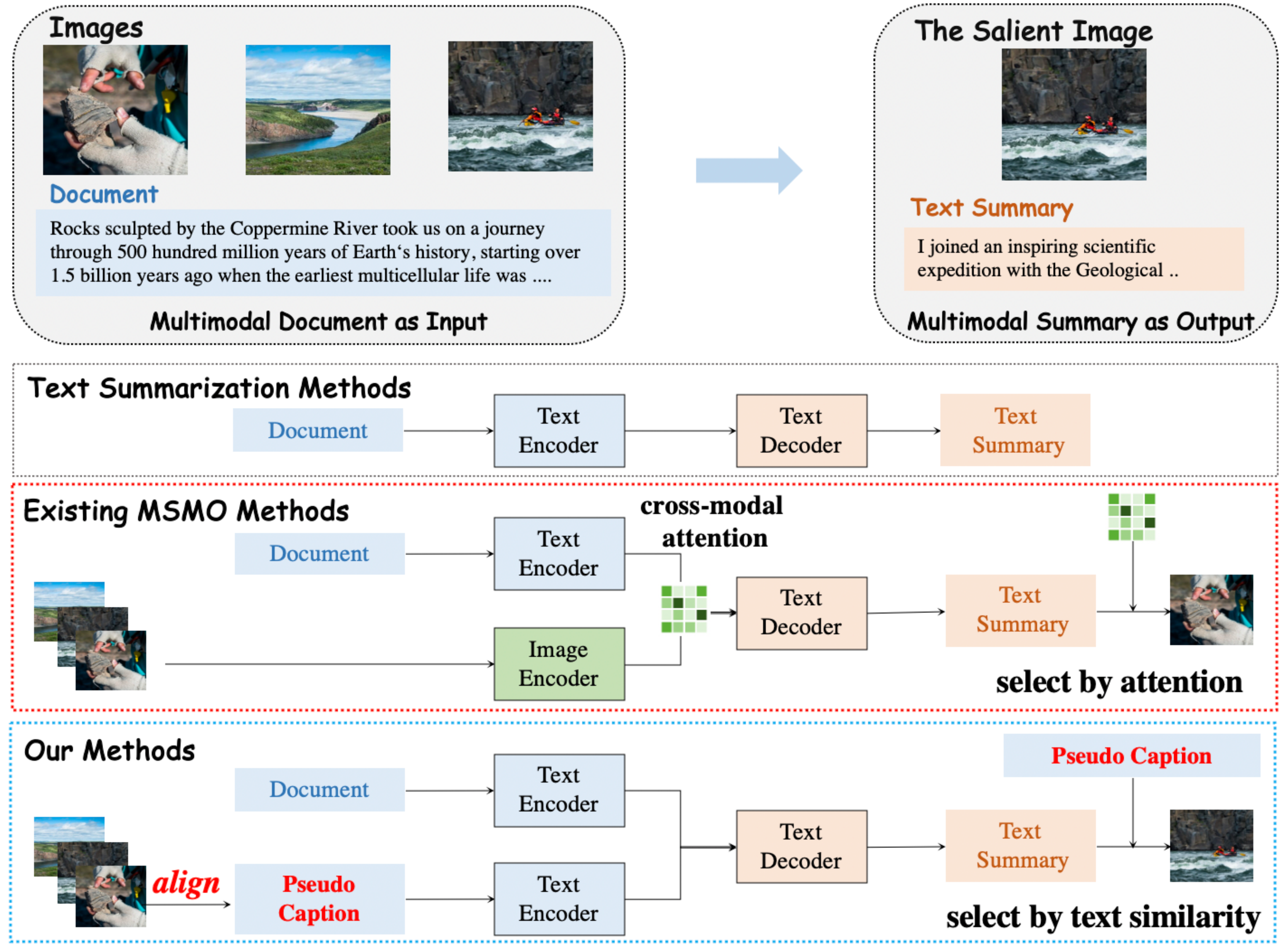}
\caption{Overview of text summarization and MSMO. Compared with text summarization models, existing MSMO methods usually use an extra image encoder to project images into intermediate representations. They identify the salient image by cross-modal attention, which could be inaccurate due to the lack of golden images for training. We explicitly transform an image into a concrete caption by image-text alignment, capturing visual knowledge better and making text summarization and image selection more effective yet simpler.}
\label{fig:MSMO}
\end{figure}

\noindent With the increase of multimedia data on the Web, multimodal summarization has drawn widespread attention from researchers in the communities of Web technologies\cite{DeepQAMVS,CCS-MMS}, natural language processing (NLP) \cite{uzzaman2011multimodal,li2017multi,li2020vmsmo, Jiang2023VisionLP} and computer vision (CV)  \cite{chen2018abstractive,palaskar2019multimodal,li2020aspect,liu2020multistage}. 
More recently, many efforts \cite{zhu2018msmo,zhu2020multimodal,Zhang2021UniMSAU} have been dedicated to multimodal summarization with multimodal output (MSMO), the novel task of generating pictorial summaries given a Web document consisting of plain text and a collection of images. As shown in Figure \ref{fig:MSMO}, a pictorial summary generated by MSMO models consists of a text summary and a salient image, delivering more user-friendly information than single-modal text summaries, according to human judgments \cite{zhu2018msmo,zhu2020multimodal}. 

MSMO faces two main challenges. (1) \textbf{There are no recommended image references available for training MSMO models.} Due to the lack of supervision signals from visual modality, it is nontrivial to optimize the cross-modal attention between texts and images, which is highly relied on by existing MSMO methods to pick salient images. According to previous best results\cite{Zhang2021UniMSAU}, only about 60\% of the predicted images are correct, indicating that image selection remains a bottleneck. 
(2) \textbf{Visual knowledge is commonly underutilized to improve text summaries.} Existing MSMO efforts show no evident improvement or even negative impact on text summaries  (e.g., decreased ROUGE scores) over typical single-modal text summarization methods. Previous literature\cite{zhu2018msmo} explained that some images were noises and long text had contained enough information for text generation, while we conjecture that these methods may not sufficiently exploit visual knowledge to characterize salient text.

To summarize, previous efforts typically encode images and texts into the same semantic space, struggling with optimizing cross-modal interaction without training signals for image selection, as the red box in Figure \ref{fig:MSMO} shows. In this dilemma, image captions, which naturally connect images and texts, can provide a cross-modal semantic bridge. Indeed, our preliminary experiments show the efficacy of introducing imageIn captions (see Section \ref{qualitiy-of-aligned-sentences}). Yet, exposure of image captions during training is inconsistent with MSMO's task settings, since MSMO excludes them to pursue better generalization of cross-modal semantic modeling\cite{zhu2018msmo,zhu2020multimodal}.On the other hand, however, it inspires us to identify a highly-relevant sentence for an image as its pseudo yet meaningful caption, providing us with a new perspective to improve MSMO. As shown in the blue box in Figure \ref{fig:MSMO}, \textit{unlike current works that represent an image as an intermediate state, we transform it into a concrete sentence to better capture visual knowledge under MSMO settings}. This transformation presents an opportunity to incorporate pre-trained visual-language models more smoothly, while making further text summarization and image selection extremely simple.

Aligning a sentence with an image could be straightforward, but identifying sentences benefiting MSMO the most is non-trivial. The reasons are two-fold. (1) \textbf{A sentence well aligned with an individual image can not guarantee a suitable one for MSMO}. An intuitive way to select a sentence is to simply retrieve it from the document, with the image as the query of a pre-trained cross-modal retrieval model. Unfortunately, we find this manner yields unsatisfactory MSMO performance  (see Section \ref{Cross-modal-Retrieval}). (2) \textbf{A classical single-pass one-to-one alignment strategy may miss salient sentences for summarization} (see Section \ref{coarse-to-fine}). There can be one-to-many and many-to-one relationships between images and sentences, and images can be similar in a document, so we need to synthesize yet distinguish image semantics from a global perspective to make better MSMO-oriented alignment.

To this end, we design a coarse-to-fine image-text alignment mechanism to produce pseudo image captions for MSMO. Firstly, a reference caption for an image is retrieved with a cross-modal retrieval model from the golden summary, rather than the whole document (Section \ref{caption-retrieval}), to capture more summary-friendly information. Since no golden summary exists at inference time, these reference captions are used to train a two-pass image-text alignment model (Section \ref{ita}) that yields pseudo captions when making inferences (that's why ``reference captions'' are so named). Given a document with ten images, for example, we will first synthesize them as a whole to select ten sentences with many-to-many coarse-grained alignment, and then identify ten individual one-to-one fine-grained matchings by bipartite graph matching over the cross-modal attention matrix. 

The pseudo image captions that imply visual knowledge are used as extra highlighted features for text summarization (Section \ref{summarization}), and the salient image is picked based on the ROUGE score between its pseudo captions and the generated summary (Section \ref{image-selection}). Extensive experiments on an existing MSMO dataset not only verify the superiority of our method but also reveal the inner connection between image captions and summaries, demonstrating promising research opportunities for our novel perspective of bridging the cross-modal semantic gap by generating pseudo image captions. 

\begin{figure*}[htpb]
\setlength{\belowcaptionskip}{-0.15cm}
\centering
\includegraphics[width=0.9\textwidth]{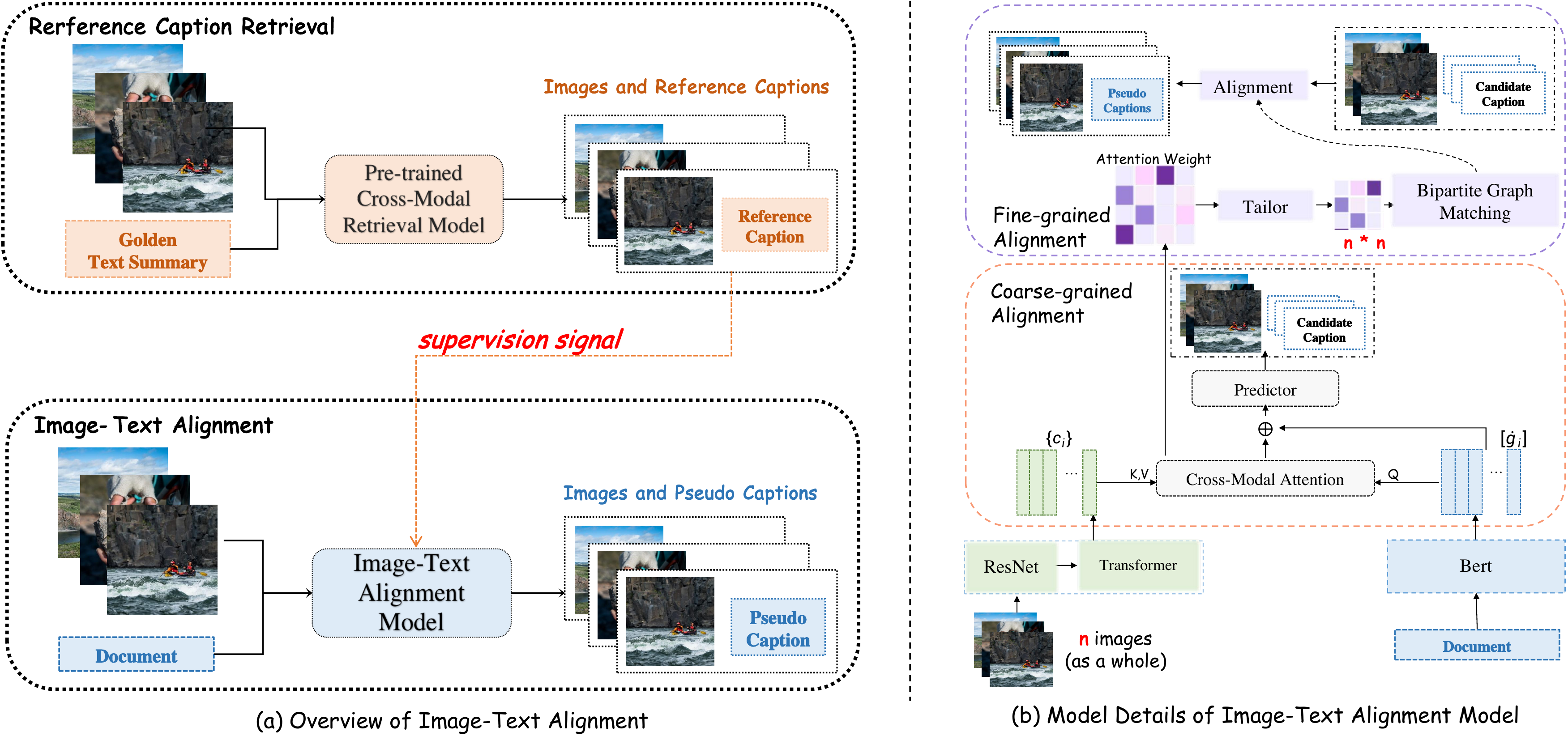}

\caption{Coarse-to-fine Image-Text alignment. The left part (figure a) shows the overview of the whole image-text alignment mechanism. Reference captions are first retrieved from golden summaries based on a cross-modal retrieval model. We then train an image-text alignment model with reference captions as supervision signals, identifying a relevant sentence as a pseudo caption for each image. The right part (figure b) demonstrates how our two-pass coarse-to-fine alignment model works internally. 
}
\label{fig2}
\end{figure*}

\section{Method}

\subsection{Problem Formulation}
~\\For MSMO task, the input is a multimodal document \{$T,V$\} including a text document $T$ with $m$ sequential sentences, where $T=[t_1,t_2,\cdots,t_m]$, and a image collection $V$ with $n$ images, where $V=\{v_1,v_2,\cdots,v_n\}$. The output is a multimodal summary $\{S,\hat{v}\}$ where $S=[s_1,s_2,\cdots,s_l]$ is a text summary containing $l$ generated sentences and $\hat{v}$ is the image selected from $V$.

\subsection{Method Overview} 
\noindent 

Our method, named \textbf{SITA}, refers to a multimodal \textbf{S}ummarization model based on a coarse-to-fine \textbf{I}mage-\textbf{T}ext \textbf{A}lignment mechanism. SITA consists of four modules: (1) ~\textbf{Reference Caption Retrieval},  (2) \textbf{Image-Text Alignment}, (3)  \textbf{Text Summarization}, and (4) \textbf{Image Selection}.
See more illustrative details in Figure~\ref{fig2} (a).

\subsection{Reference Caption Retrieval}
\label{caption-retrieval}

\noindent Given the multimodal document $\{T,V\}$, we first retrieve reference captions from the golden text summary for each image in $V$, based on a pretrained cross-modal retrieval model consisting of an image encoder and a text encoder. The image encoder is ResNet152 \cite{he2016deep} pretrained on ImageNet \cite{deng2009imagenet} and the text encoder is a BERT-based sentence encoder for text summarization \cite{liu2019text}. Following \cite{Faghri2018VSEIV}, we train the model on the COCO dataset \cite{lin2014microsoft} by matching image representations and sentence representations.

We retrieve reference image captions from the golden summary rather than the whole document, to make the retrieval results more summary-friendly and narrower-focused (see Section\ref{Cross-modal-Retrieval}). However, a new dilemma is the lack of golden summaries during inference.
Therefore, we exploit them to train an image-text alignment model, which predicts pseudo captions during inference.

\subsection{ Coarse-to-fine Image-Text Alignment}
\label{ita}
\noindent 
\noindent We design a coarse-to-fine \textbf{I}mage-\textbf{T}ext \textbf{A}lignment model (\textbf{ITA}) with training signals obtained from reference captions, to generate pseudo image captions. 
Since there can be one-to-many and many-to-one relationships between images and sentences, employing a simple single-pass one-to-one alignment strategy tends to generate a limited set of aligned sentences
repeatedly, incapable of recalling enough
relevant sentences (see Section \ref{coarse-to-fine}). To this end, we propose a novel two-pass coarse-to-fine mechanism to align sentences better.

Specifically, for the $n$ images in $V$, we will take them as a whole to select $n$ sentences from the document $T$ with coarse-grained alignment, and then identify one-to-one matchings via fine-grained alignment. ITA consists of an image encoder, a sentence encoder, a coarse-grained alignment module, and a fine-grained alignment module. 

\subsubsection{Image Encoder}

We first use ResNet152 to extract image features for each image in $\{v_1,v_2,\cdots,v_n\}$. These features are then fed into a Transformer-based encoder \cite{2017Attention} as a whole to synthesize global knowledge and interaction information among all images. The position embeddings are not used here since image order information is unavailable. The final output of the image encoder is denoted as $\{c_1,c_2,\cdots,c_n\}$.

\subsubsection{Sentence Encoder} 

The sentence encoder here is the same as the one used in reference caption retrieval. For $m$ sentences denoted as $[t_1,t_2,\cdots,t_m]$, the sentence encoder generate a representation sequence $[g_1,g_2,\cdots,g_m]$.

\subsubsection{Coarse-grained Alignment} 

To do coarse-grained alignment, we first apply a cross attention between sentences and images to refine sentence representations:

\begin{equation} o_{i,j} = \frac{Q_i\cdot K_j}{\sum^{n}_{k=1}Q_i\cdot K_k} \end{equation}
\begin{equation} a_{i,j} =\frac{exp({o_{i,j}})}{\sum^{n}_{k=1}exp({o_{i,k}})} \end{equation}
\begin{equation} \dot{g}_i=Q_i +\sum_{k=1}^{n}a_{i,k}\times V_k \end{equation} where $Q_i = W_q\times g_i $, $K_j = W_k \times c_j$, $V_j = W_v \times c_j$, $ i \in \{1,2,\cdots, m\}$, and $j\in \{1,2,\cdots, n\}$. $W_q, W_k, W_v \in R^{D\times D}$ are learnable parameters, where $D$ is 768 (the dimension of the image/text feature vectors). Noted that we have calculated an attention matrix $A \in R^{m \times n}$ based on the equation 1 and 2 where $a_{i,j} $ is the element in the i-th row and j-th column of $A$. 

The refined representation $\dot{g}_i$ is then fed to a sigmoid classifier to predict whether sentence $t_i$ will be selected:
\begin{equation} p_i = \sigma(W_p~\dot{g}_i+b)\end{equation}
where $W_p\in R^{D \times D}$ and $b \in R^{D}$ are learnable parameters.


To train the model, we need $n$ recommended sentences as references for a multimodal document with $n$ images. For each image $v_i$, we will calculate the ROUGE scores between sentences in the document and their reference captions generated in the first step, and the sentence with the highest score will be labeled as selected. If a sentence is selected more than once, we will pick another sentence with the next highest score. We use $y_i=1$ to denote that sentence $t_i$ is selected, and $y_i=0$ otherwise. Then, for the $m$ sentences in the document $T=[t_1,t_2,\cdots,t_m]$, we employ the binary cross-entropy loss to optimize the model as follow:
 \begin{equation} \mathcal{L}_{BCE} = - \frac{1}{m}\sum^{m}_{i=1} y_ilog(p_i)+(1-y_i)log(1-p_i)\end{equation}

 \subsubsection{Fine-grained Alignment} 


Based on the coarse-grained alignment, we have calculated the an ${m \times n}$ cross-modal attention matrix (denoted as $A$), in which the element in the $i$-th row and $j$-th column is $a_{i,j}$. In this step, we want further to identify optimal one-to-one relationships between images and these sentences. Generally, the larger the attention weight between $t_i$ and $v_j$, the more likely $t_i$ and $v_j$ match.
 Suppose we have obtained $n$ selected sentences denoted as $t_{z_1},t_{z_2},\dots,t_{z_n}$ and we extract the rows corresponding to these sentences from the matrix $A$ and concatenate them as a new attention matrix $\dot{A}$ : 
\begin{equation}
    \dot{A} = concat([A_{z_1},A_{z_2},\dots,A_{z_n}])
\end{equation} 
where $\dot{A}\in N^{n\times n}$, $A_{z_i}\in R^{n}, i\in\{1,2,\dots,n\}$.
Based on the new cross-modal attention matrix $\dot{A}$, we can construct a complete weighted bipartite graph $G$ containing two disjoint and independent vertice sets $S$ and $V$, where $\left| S\right|=n$ and $\left|V\right|=n$. So there are $n \times n$ weighted edges in $G$. The vertice $v_i$ in $V$ represents an image, and vertice $s_j$ in $S$ represents a sentence. The weight of the edge in $G$ between the vertice $v_i \in V$ and the the vertice $s_j \in S$ is the value $a_{ij} \in R$ in $\dot{A}$. Therefore, the fine-grained alignment of the sentences and images can be regarded as a maximum-weight perfect matching in the bipartite graphs $G$. We can easily utilize the bipartite graph matching algorithm (Kuhn-Munkres algorithm \cite{2010The} in our implementation) to match the vertices in the two sets in the graph:
\begin{equation}
    M = KM(\dot{A}) 
\end{equation}

\noindent where $M = \left[I_1,I_2,\dots,I_n\right], I_i \in \{1,2,\dots,n\}$ represents the index list of selected sentences(e.g., the first image is aligned with the $I_1$-th sentence in the selected sentences), and KM represents the Kuhn-Munkres algorithm. 


\subsection{Text Summarization}
\label{summarization}

\noindent We build the text summarization module based on {BERTSum}, a recent simple yet robust summarization model \cite{liu2019text}. 
We concatenate all pseudo image captions as a new text document denoted as $T_s$. The origin text document $T$ and the new text document $T_s$ are fed into the encoder of {BERTSum} separately, generating two representation sequences $R$ and $R_s$. Then, unlike the traditional Transformer decoder, 
we have two individual cross attention modules—corresponding to the two documents—after the self-attention module in each Transformer block. The outputs of the two cross attention modules are simply summed, leaving other components in the Transformer block unchanged. 


\subsection{ Image Selection}
\label{image-selection}
 \noindent Given the generated summary denoted as $S$ and pseudo captions $\{t_{z_1}, t_{z_2}, \dots, t_{z_n}\}$, the image $\hat{v}$ whose pseudo caption $\hat{t}$ generates the highest ROUGE-L with the summary $S$, is selected as the most salient image, where:
\begin{equation}
    \hat{t}  = \mathop{\arg\max}\limits_{t_k} (R(t_k,S))
\end{equation}
 $k \in \{z_1,z_2,\dots,z_n\}$ and $R(t_k,S)$ represent the function which calculates the ROUGE-L socre between $t_k$ and $S$.

Please refer to appendix \ref{implementation} and our released code for more architecture and implementation details$\textsuperscript{\ref{gitlink}}$.

\section{Experiment Settings}

\subsection{Dataset}

\noindent We use the dataset build by \citet{zhu2018msmo}, which is constructed from the Daily Mail website$\footnote{http://www.dailymail.co.uk/}$,
and contains 293,965 articles for training, 10,355 articles for validation, and 10,261 articles for testing. Please refer to appendix \ref{dataset} for more dataset details.

\begin{table*}[htbp]
\centering
 
\setlength{\tabcolsep}{4.3mm}{

\begin{tabular}{llllcccc}
\toprule[1.3pt]

 Model   & R-1            & R-2            & R-L            &   IP     & M$_{sim}$   & MR$_{max}$ &    MMAE++   \\ \hline

ATG & 40.63          & 18.12          & 37.53          & 59.28          & 25.82 &     56.54 &    67.63  \\
ATL & 40.86          & 18.27          & 37.75          & 62.44          & 13.26 &     55.67 &    67.26   \\
HAN & 40.82          & 18.30          & 37.70          & 61.83          & 12.22 &     55.29 &    66.93\\
 
MOF & 41.20          & 18.33       & 37.80          &  65.45          & 26.38 &    58.38 &   69.66  
\\
UniMS & 42.94 & 20.50 & 40.96 &  69.38 & 29.72 & - & - 
\\\hline
 
SITA (Ours) & \textbf{43.64}  & \textbf{20.53} & \textbf{41.03} & \textbf{76.41} & \textbf{33.47}  &   \textbf{65.38} &   \textbf{77.91}  \\
\bottomrule[1.3pt]
\end{tabular}}
\caption{Main results of different metrics. R-$\{1,2,L\}$ refers to ROUGE-$\{1,2,L\}$. }
\vspace{-0.4cm}
\label{table1}
 
\end{table*}
\begin{table}[htpb]
\centering
\small
\setlength{\tabcolsep}{1.7mm}{
\begin{tabular}{llll}

\toprule[1.3pt]
Model              & R-1            & R-2            & R-L           \\ \hline
PGN & 41.11          & 18.31         & 37.74         \\\hdashline
ATL & 40.86($\downarrow $0.05)   & 18.27($\downarrow$0.04)  & 37.75($\downarrow$0.01) \\
MOF & 41.20($\uparrow $0.09)    & 18.33($\uparrow$0.02)       & 37.80($\uparrow$0.06) \\
\hline
BERTSum     &   41.51        &   19.43     & 38.85  \\\hdashline
SITA& \textbf{43.64}($\uparrow$2.13) & \textbf{20.53}($\uparrow$1.10) & \textbf{41.03}($\uparrow$2.18) \\  
\hline
BART  & 41.83 &  19.83 &  39.74 \\\hdashline
UniMS & \textbf{42.94}($\uparrow$1.11) & \textbf{20.50}($\uparrow$0.67) & \textbf{40.96}($\uparrow$1.22) \\ 
 \bottomrule[1.3pt]
\end{tabular}}
\caption{Comparison of text summary quality in terms of Rouge scores. The numbers in parentheses represent relative performance improvements of multimodal models over their single-modal ones (e.g., PGN, BERTSum and BART). $\uparrow$ indicates a positive effect, and $\downarrow$ indicates a performance decrement.}
\label{table_text_sum}
\vspace{-0.4cm}
\end{table}

\subsection{ Evaluation Metrics}
\noindent Following \citet{zhu2018msmo,zhu2020multimodal}, we choose the following metrics.
(1) \textbf{ROUGE-$\{1,2,L\}$} is the standard text summarization evaluation metric.
(2) \textbf{IP} is the abbreviation of \textbf{I}mage \textbf{P}recision and used to evaluate image selection. It is defined by dividing the size of the intersection between the recommended images $rec_{img}$ and the reference images $ref_{img}$ by the number of recommended images.
(3) \textbf{M$_{sim}$} evaluates the image-text relevance by calculating the maximum similarity between the image and each sentence in the model summary.
(4) \textbf{MR$_{max}$} evaluates the information integrity of the multimodal summary. It exploits a joint multimodal representation to calculate the similarity between model outputs and multimodal references.
(5) \textbf{MMAE++} evaluates the overall quality of multimodal summaries.  It projects both the candidate multimodal summary and the reference summary into a joint semantic space with a trained neural network. 
For the details of MMAE++, please check subsection 3.3 in \citet{zhu2018msmo}'s work.

Meanwhile, we propose \textbf{Caption-ROUGE-$L$}, a metric specific to SITA and its variants by calculating ROUGE-$L$ between a generated pseudo caption and the golden caption.

\subsection{ Baselines}
\noindent We compare our method with the five multimodal summarization methods. 
(1) \textbf{ATG} \cite{zhu2018msmo} is a multimodal attention model, which measures image salience by the visual attention distribution over the global image features.
(2) \textbf{ATL} is an  ATG variant using attention distributions over image patches.
(3) \textbf{HAN} is an ATL variant by adding a hierarchical attention mechanism on image patches.
(4) \textbf{MOF} \cite{zhu2020multimodal} introduces a multimodal objective function into ATG.
Among the four MOF variants, we choose the one having the best performance in five of the seven metrics we used. 
(5) \textbf{UniMS} \cite{Zhang2021UniMSAU} is a recent unified framework for multimodal summarization.


We also compare our method with the three text summarization methods.
(1) \textbf{PGN} \cite{see2017get} is the Pointer-Generator Network for abstractive text summarization model. 
(2) \textbf{BERTSum} is a recent robust BERT-based summarization model proposed by \citet{liu2019text}, upon which our SITA is built.
(3) \textbf{BART} \cite{Lewis2020BARTDS} is a pretrained seq2seq model consisting of a bidirectional encoder and an auto-regressive decoder.

\section{Experiment Results}

\subsection{ Main Results}


\noindent Table~\ref{table1} and ~\ref{table_text_sum} show the performance of the baseline models and our method. By investigating the results, we have the following observations.	

(1) Our SITA achieves improvements over baselines across all evaluation metrics of image precision, text summary quality, image-text relevance, and multimodal information integrity, clearly setting up a new state-of-the-art performance.

(2) Regarding the visual modality metric (IP), MOF generally outperforms its predecessor baselines by a slight margin due to its auxiliary training objective of image selection. UniMS further gain a notable improvement over MOF by distilling knowledge in a vision-language pre-trained model. \textit{Our SITA impressively improves more than 10\% over UniMS in the precision of recommended images (e.g., 76.41 of SITA v.s. 69.38 of UniMS on the IP metric).} The reason is that the pseudo captions identified by our coarse-to-fine alignment mechanism provide much more informative clues for image selection. We will provide more detailed analyses in the following experiments.

(3) Regarding textual modality metrics, more comprehensive comparisons are shown in Table \ref{table_text_sum}, which consists of three groups of results. In the first group, existing multimodal methods (ATL and MOF) demonstrate no superiority over the single-modal text summarization model they used  (PGN). These efforts concluded that too many images could bring noise, and the long document had contained enough information for text generation \cite{zhu2018msmo,zhu2020multimodal}. In contrast, our SITA (in the second group) gains a much more remarkable improvement, e.g., of 2.18 ROUGE-$L$, on text summaries, even based on a more robust base model (BERTSum). The latest state-of-the-art UmiMS (in the third group), built upon BART, also achieves performance improvements (e.g., +1.22 ROUGE-$L$) on text summarization, but not as evident as ours. Note that BART performs better than BERTSum on text summarization (e.g., 39.74 v.s. 38.85 of ROUGE-$L$), but SITA still outperforms UmiMS. \textit{These results suggest that visual information actually benefits text generation, and our method exploits it more effectively}.

(4) M$_{sim}$, MR$_{max}$, and MMAE++ are used to check image-text relevance, image-text integrity, and the overall effectiveness of pictorial summaries. As expected, SITA maintains dominance over baselines on the three intermodality metrics. These superiorities come from remarkable improvements on intramodality metrics and SITA's inherent capabilities of bridging the cross-modal semantic gap. 

Note that IP and all intermodality metrics depend on the selected salient images, hence indirectly relying on the generated text summaries. Rigorously,  baseline methods and our SITA utilize different text summarization models (e.g., PGC, BART, and BERTSum), so these metrics will be more friendly to methods with better-performed base text summarization model. However, this fact has minor impacts on our above analyses, since image selection improvements of SITA mainly benefit from pseudo captions but not the text summaries.

\subsection{ Effects of the Coarse-to-fine Mechanism}
\label{coarse-to-fine}
\begin{table}[htpb]
\small
\centering
 \setlength{\tabcolsep}{1.3mm}{
\begin{tabular}{lccccc}
\toprule[1.3pt]
Model                & R-1              & R-2            & R-L            & IP     & CR$-L$ \\ \hline
SITA     & \textbf{43.64} & \textbf{20.53} & \textbf{41.03} & \textbf{76.41}& \textbf{39.39} \\
  ~~-\textit{w/o} ITA  &   41.79          & 19.54          & 38.97          &   72.95      &     38.23       \\
\hline  
 One-pass  &   40.83          & 18.32          & 37.98          &   57.28      &     12.31 \\
 One-pass(Dedup)  &   41.67 & 18.98 & 38.63 & 64.32 & 33.21       \\

 \bottomrule[1.3pt]
\end{tabular}}
\caption{Performance of SITA and its variants. CR$-L$ refers to Caption-ROUGE-$L$. -\textit{w/o} ITA directly retrieves pseudo captions in a document without image-text alignment, One-pass does image-text alignment in a single-pass manner, and One-pass (Dedup) adds an sentence deduplication mechanism over One-pass. }
\label{table-cross-modal-retrieval}
\vspace{-0.5cm}
\end{table}

\subsubsection{Results of One-pass Alignment Strategy.}

To investigate how the coarse-to-fine alignment strategy boosts performance,  we replace it with a single-pass alignment method, which is trained to select a pseudo caption for only one single image at a time. The results of this method variant (named \textbf{One-pass}) are shown in Table ~\ref{table-cross-modal-retrieval}, from which we see notable performance degradation on all metrics. Through further qualitative exploration on its prediction results, we find this method tends to generate a small set of sentences repeatedly among different images, incapable of recalling enough relevant sentences. The low Caption-ROUGE-$L$ of \textbf{One-pass} (e.g., 12.31) also verifies this observation. One possible reason is that images in a document can sometimes be similar, making the single-pass strategy fail to characterize the correlation and difference among these images. \textit{In contrast, by introducing the coarse-to-fine mechanism, our alignment model synthesizes multiple images from a global perspective in the coarse-grained pass, recalling more sentences more accurately and hence facilitating further fine-grained alignment}.

\subsubsection{Comparison with Simple Deduplication}

To avoid recalling repeated sentences in one-pass alignment, one simple alternative strategy is introducing a deduplication mechanism. We hence implement \textbf{One-pass (Dedup)}, which will select another sentence with the next highest score if the current sentence has been chosen. As shown in Table ~\ref{table-cross-modal-retrieval}, we can see that the deduplication mechanism over one-pass image text alignment brings improvements (e.g., +0.65 on R-L and +7.04 on IP). But the performance of \textbf{One-pass (Dedup)} is still far from our full SITA with the coarse-to-fine alignment strategy (e.g., with a significant gap of 2.4 on R-L and 12.09 on IP). The main reason is that one image may align with multiple semantically rich sentences. For such an image, even with the deduplicating mechanism, one-pass alignment can only recall a single sentence, potentially missing critical information, especially when other images do not semantically overlap with it. That roughly explains the performance gaps. This comparison further verifies the necessity and soundness of the technical design of the two-pass coarse-to-fine alignment.

\begin{figure}[htbp]
\centering
\includegraphics[width=0.96\linewidth]{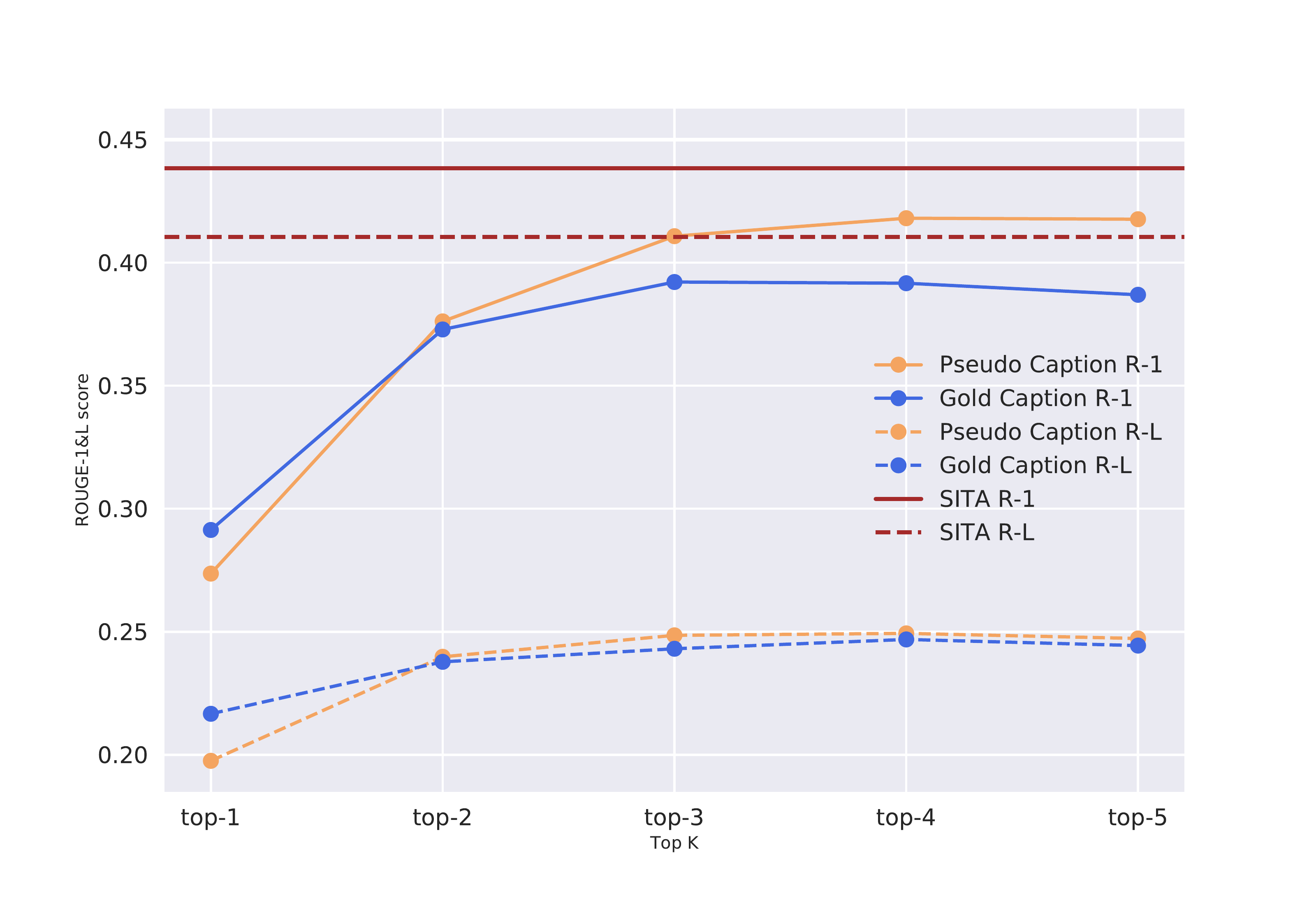}


\caption{ROUGE-1 and ROUGE-L scores of simple summaries generated by simply concatenating pseudo captions (orange) or golden captions (blue) of a document's first $k$ images. The scores are calculated by matching them against the reference summaries. The horizontal red (dashed) lines represent the text summaries generated by SITA. ROUGE-2 is similar to Rouge-1, which is not shown for better visualization.}
\label{fig_simple_summary}
 \vspace{-0.2cm}
\end{figure}

\subsection{ Effects of Cross-modal Retrieval}
\label{Cross-modal-Retrieval}

\noindent 
To investigate the effect of the cross-modal retrieval, we directly retrieve pseudo captions in a document (rather than a summary), obtaining another method variant (named \textbf{\textit{w/o} ITA}) requiring no image-text alignment training anymore.

As shown in Table ~\ref{table-cross-modal-retrieval}, \textbf{\textit{w/o} ITA} bring modest enhancement on text summaries (e.g., 38.85 of BERTSum v.s. 38.97 of \textbf{\textit{w/o} ITA} on ROUGE-$L$), while achieving more impressive image salience (e.g., 72.28 on IP). Compared with our full SITA, this method variant (named \textbf{\textit{w/o} ITA}) demonstrates significant performance degeneration on both text and image salience (e.g., -1.06 on ROUGE-$L$ and -3.04 on IP). These results reveal that \textit{(1) the knowledge in the pre-trained cross-modal retrieval model mainly helps image selection, and the image-text alignment over retrieval results is more critical for the overall performance; and (2) retrieving reference captions from summaries instead of documents is a key design of SITA.}

Note that our cross-modal retrieval model is pre-trained with  113K image-text pairs. Though UniMS distills knowledge from a vision-language model pre-trained by more than 400M image-text pairs, SITA demonstrates significant superiority.

\subsection{Quality of Pseudo Captions }
\label{qualitiy-of-aligned-sentences}

\begin{table}[htpb]
\centering
\small
\setlength{\tabcolsep}{1.9mm}{
\begin{tabular}{cccccc}

\toprule[1pt]
Model              & R-1            & R-2            & R-L            & IP    & CR$-L$ \\ \hline
Caption-train & 42.22          & 19.70          & 39.29          & 73.59 & \textbf{39.54}  \\
Caption-input & 42.71          & 20.04          & 39.85           & 75.33 & -      \\
SITA    & \textbf{43.64} & \textbf{20.53} & \textbf{41.03} & \textbf{76.41} & 39.39  \\ \bottomrule[1pt]
\end{tabular}}
\caption{Comparisons of SITA with models using golden captions. Caption-train use golden captions to train image-text alignment model, and Caption-input directly use golden captions as input for text summarization and image selection. CR$-L$ refers to Caption-ROUGE-$L$.  Caption-input does not generate pseudo captions, so its CR-$L$ is unavailable.}
\label{table-ablation} 
\vspace{-0.3cm}
\end{table}

 \noindent We further analyze the effectiveness of our method from the perspective of pseudo captions' quality. We are interested in the relation between golden captions and our pseudo captions. In the MSMO's task settings, golden image captions are excluded.
 To perform this study, we allow the compared models to use golden captions in training under a easier task setting. Here we build another two baselines. 
 
The first one, named ~\textbf{Caption-train}, \textbf{train}s the image-text alignment model with golden captions instead of the reference sentences retrieved in the first step. We compare SITA with it on the metrics of ROUGE-$\{1,2, L\}$, IP, and Caption-ROUGE-$L$. Looking into the empirical results shown in Table \ref{table-ablation}, the Caption-ROUGE-$L$ of SITA and \textbf{Caption-train} are generally similar. Hence, from the perspective of recovering image captions, the quality of aligned sentences generated by \textbf{Caption-train} and SITA are identical. However, SITA generates better text summaries and salient images than \textbf{Caption-train  } (e.g., with improvements of  0.74 on ROUGE-$L$ and 2.82 on IP), suggesting that our aligned sentences benefit more MSMO. The reason is that the \textit{reference captions used for alignment training are retrieved from text summaries, inherently making predicted pseudo captions imply better summary features.}

The second one, named ~\textbf{Caption-input}, directly utilizes golden captions instead of pseudo captions as \textbf{input}s for text summarization and image selection. We find that SITA also outperforms \textbf{Caption-input} on all metrics. The performance enhancement is less evident but still impressive, considering that SITA uses a more restricted task setting. This observation proves that \textit{the pseudo captions we generated are even better than the original image captions for MSMO.}

The above analyses verify that pseudo captions are not only semantically consistent with images but also informative for text summarization. 

\subsection{Correlation between Image captions and Text Summaries}


We also investigate the correlation between image captions and text summaries. 
Specifically, we construct a simple summary by concatenating golden (or pseudo) captions of the first $k$ images in a document. Then, we calculate the ROUGE scores of those simple summaries. The results are shown in Figure \ref{fig_simple_summary}, and we have the following observations:

(1) Simply aggregating some (pseudo) image captions can generate generally good summaries. For example, when selecting more than three captions, the resulting summaries even have a better ROUGE-1 than MOF. The observation verifies the inherent capabilities of image captions on the briding cross-modal semantic gap.

(2) The upward trend of the ROUGE-$L$ with the increase of $k$ is not as notable as that of ROUGE-1. The reason is that text generated by sentence concatenation (in random order) may lack coherence. ROUGE-$L$ is calculated based on the longest common substring, the length of which will be limited in this situation. This phenomenon suggests that an individual text summarization component is still required given these high-quality image captions. 

(3) Generally, the red line is above the blue line most of the time,
indicating that \textit{simple summaries constructed by pseudo captions are even better than their counterparts consisting of golden captions.} The observation, again, verifies that pseudo captions generated by our image-text alignment mechanism are more informative than the original ones, in terms of improving MSMO performance.

\section{Related Work}
\noindent 
Existing text summarization approaches can be roughly categorized into extractive summarization \cite{narayan2018ranking,xiao2019extractive,zhong2020extractive,wang-etal-2020-heterogeneous} and abstractive summarization\cite{Syed2021ASO, paulus2018deep,zhang2020pegasus,Lewis2020BARTDS,Tan2017AbstractiveDS}. 
Classical abstractive summarizaition model such as Pointer Generator Network\cite{see2017get} and BERTSum\cite{liu2019text} serve as fundamental components for previous MSMO works.

Multimodal summarization takes data of more than one modalities as input and synthesizes information across different modalities to generate the output \cite{uzzaman2011multimodal,Li2018MultimodalSS, sanabria2018how2,Fu2020MultimodalSF,im-etal-2021-self,Yu2021VisionGG,zhu2018msmo,zhu2020multimodal,li2020vmsmo,Jangra2020TextImageVideoSG,Jangra2020MultiModalSG,jangra2021multi,Zhang2021HierarchicalCS}. 
\citet{zhu2018msmo} first propose generating pictorial summaries given a document and an image collection. \citet{zhu2020multimodal} further introduced a extra cross-entropy loss for image selection. 
Recently, \citet{Zhang2021UniMSAU}  proposed to utilize knowledge distillation with a vision-language pre-trained model to help image selection,  but the image precision was still far from ideal.

\section{ Conclusion}
\noindent We have presented SITA, a multimodal \textbf{S}ummarization method based on coarse-to-fine \textbf{I}mage-\textbf{T}ext \textbf{A}lignment.  SITA introduces a novel perspective of bridging the semantic gap between visual and textual modality by exploiting pseudo image captions. Our cross-modal alignment mechanism effectively generates high-quality pseudo image captions, enabling SITA to set up state-of-the-art performance easily. We discuss the feasibility and potential of leveraging pseudo image captions , and release code\footnote{https://github.com/sitaProject/SITA\label{gitlink}}, to inspire more future studies from our proposed perspective.

\section*{Limitations}
\noindent By retrieving pseudo captions from summaries, one limitation is that the most relevant sentence for a specific image may not be in the summary. However, it has a trivial impact on the overall MSMO performance. If this happens, most of the time, the image will not be the salient image to select, and its caption will provide no helpful information for the text summary. In this situation, selecting a pseudo caption from summary sentences will not hinder the overall performance, though it may not be the best for the specific image.

Besides, even though our task setting (including the dataset and all evaluation metrics we used) strictly follows three previous works \cite{zhu2018msmo,zhu2020multimodal,Zhang2021UniMSAU},  another possible limitation is that only one MSMO benchmark is used (no other dataset exists). We believe providing more diversified datasets and investigating more about the rationale under the task setting are critical to pushing forward the multimodal summarization community, although they are out of the scope of this work.

\bibliography{custom}
\bibliographystyle{acl_natbib}
\clearpage
\appendix

\section{Implementation Details}
\label{implementation}
We use Pytorch-Transformers$\footnote{https://pytorch.org/hub/huggingface\_pytorch-transformers/}$ to implement the Bert-base model. We use the Adam optimizer \cite{2014Adam} and set the learning rate to 0.0001. We limit the text length to 512 tokens and resize the resolution of each image to 224$\times$224. 
The overall process is implemented with PyTorch\cite{2019PyTorch}. We run our experiment using 2 NVIDIA V100 GPUs. The maximum number of training iterations is set to 200k, and we save the checkpoint every 2k iterations. We select the best checkpoints according to the validation loss and report the results on the test set. 
When training the image text alignment model, we freeze the weight of ResNet152 and use a maximum batch size of 512. 
When training the text summarization model,  we use beam search in decoding and set the beam search size to 5. The batch size is set to 512, and each input in the batch contains a text article with 512 tokens and a pseudo caption set with 128 tokens. 
For more implementation details, please refer to our released code at Github$\footnote{https://github.com/sitaProject/SITA\label{gitlink}}$.

\section{Dataset}
\label{dataset}
\begin{table}[htpb]
\small
\setlength{\tabcolsep}{1.3mm}{
\begin{tabular}{lccc}
\toprule[1.3pt]
                        & Train  & Valid  & Test   \\ \hline
\#Documents             & 293965 & 10355  & 10261  \\
\#AvgImgsNum             & 6.56   & 6.62   & 6.97   \\
\#AvgTokensNum(Document) & 720.87 & 766.08 & 730.80 \\
\#AvgTokensNum(Summary)  & 70.12  & 70.02  & 72.16  \\ \bottomrule[1.3pt]
\end{tabular}}
\caption{MSMO Dataset statistics.}
\label{msmo_dataset}
\vspace{-0.5cm}
\end{table}

\noindent We use the MSMO dataset build by \citet{zhu2018msmo}, which is the largest benchmark dataset. This dataset is constructed from the Daily Mail website$\footnote{http://www.dailymail.co.uk/}$,
containing 293,965 articles for training, 10,355 articles for validation, and 10,261 articles for testing. Each article contains a text document, and approximately seven images on average. The manually written highlights offered by Daily Mail are taken as a reference text summary. Noted that the pictorial summaries are only available on the test set, so there is no label information about the salient images during training. Image captions are excluded from the dataset for generalization.

\begin{figure}[htbp]
\centering

\centering
\includegraphics[width=0.96\linewidth]{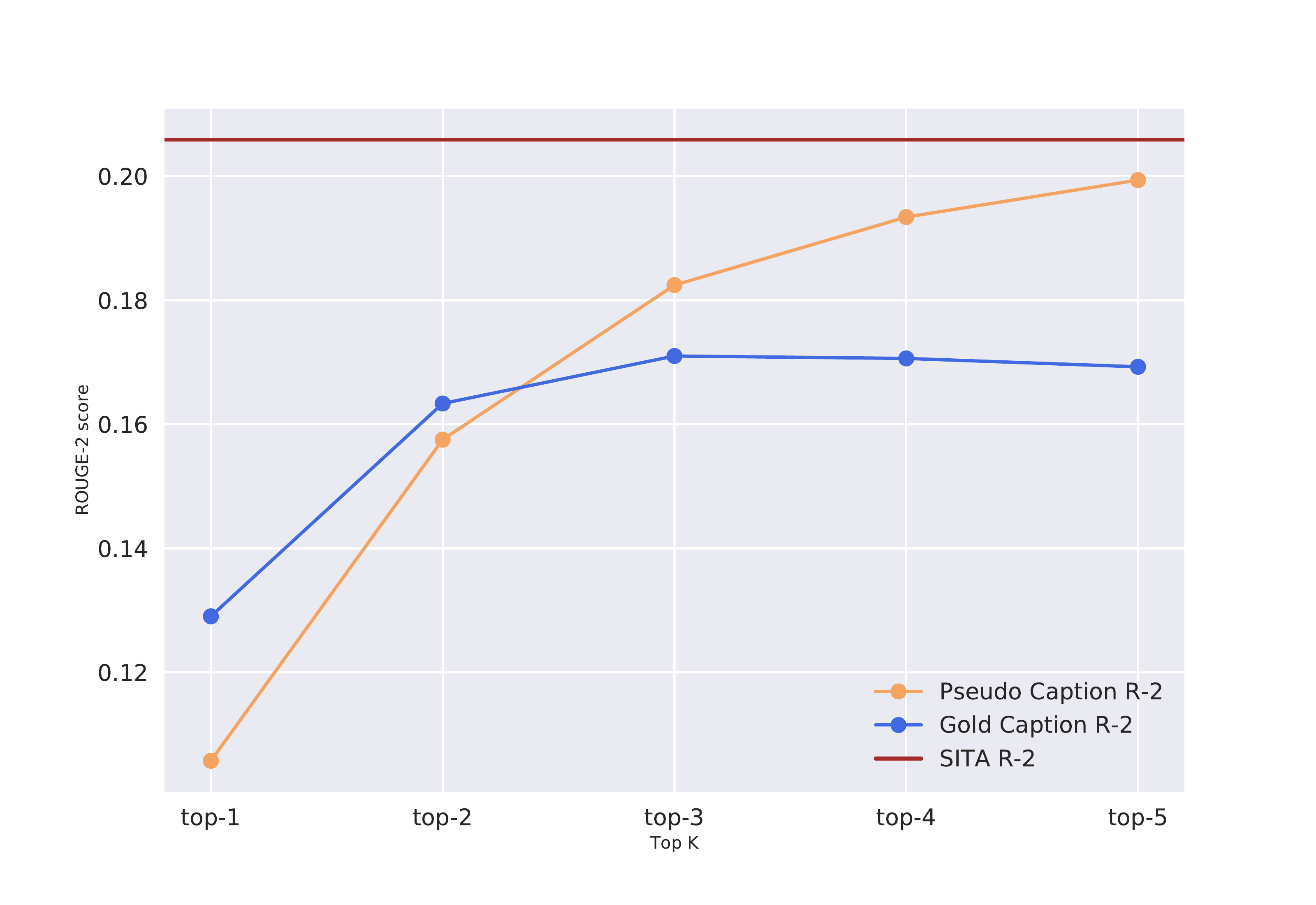}

\caption{ROUGE-2 of simple summaries generated by simply concatenating pseudo captions (red) or golden captions (blue) of a document's first $k$ images. The scores are calculated by matching them against the reference summaries. The horizontal red dashed lines represent the text summaries generated by our SITA model.}
\label{fig4}
\end{figure}

\begin{figure*}[ht]
\centering
\includegraphics[width=0.97\textwidth]{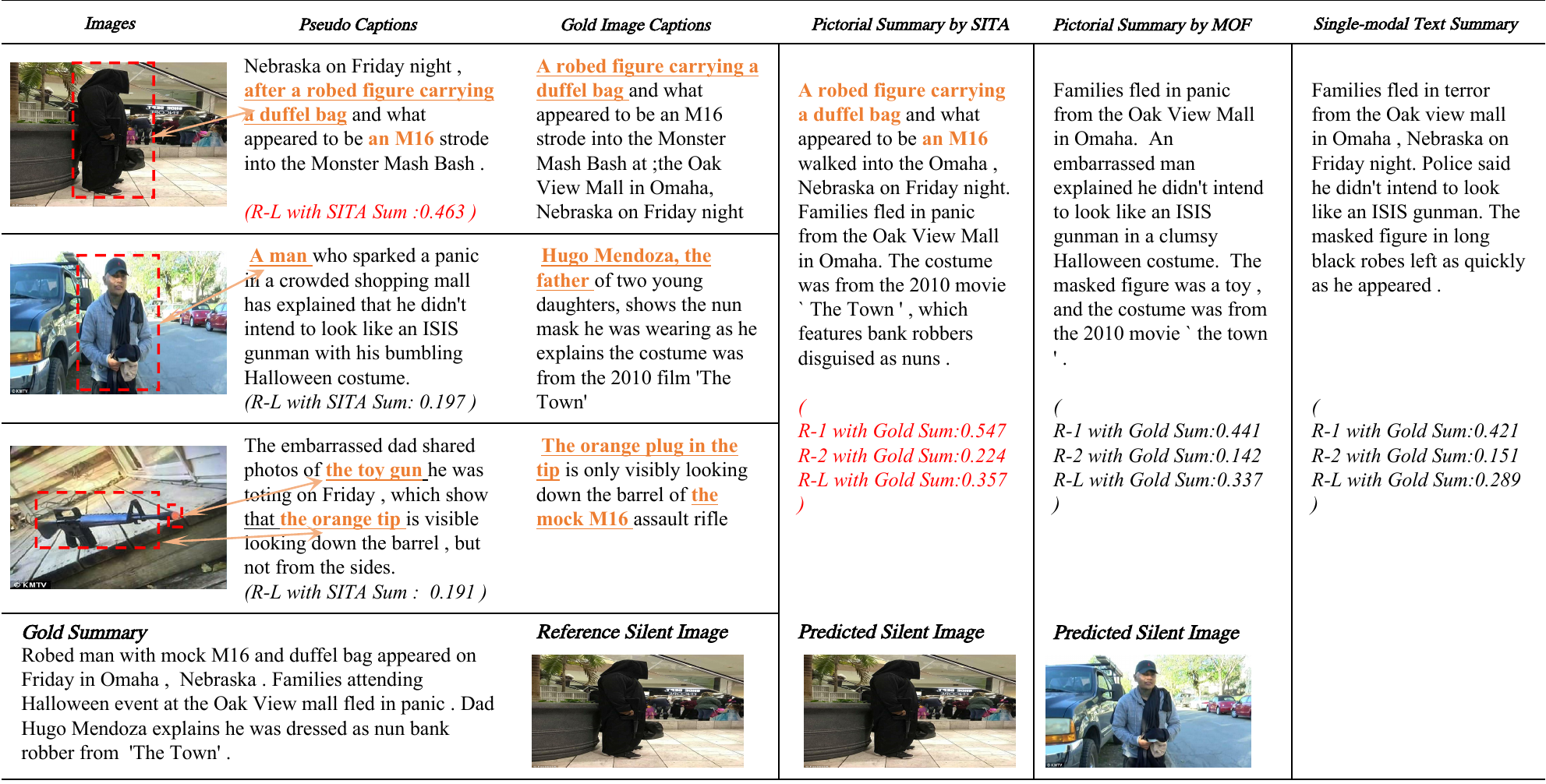} 
\caption{An example of multimodal summarization generation. The text summary in the last column is generated by the single-modal text summarization model BertSum \cite{liu2019text}. The pictorial summary in the second last column is generated by MOF \cite{zhu2020multimodal} re-implemented by ourselves. The orange text in the pseudo and gold image captions corresponds to the semantically important entities in red boxes of images.}
\label{fig9}
\end{figure*}

\section{Case Study}
To qualitatively verify our proposed method's effectiveness, we conduct a case study on generated pseudo image captions and multi-modal summaries. As illustrated in Figure ~\ref{fig9}, the pseudo captions generated by our model can express image semantics appropriately. For the critical entities in the images, we can find the corresponding descriptions in the high-quality pseudo captions we produce. Compared with the text summary generated by single-modal and alternative multi-modal models, SITA's output captures the article's main point better, thanks to the effective incorporation of pseudo image captions implying visual knowledge. For example, the descriptions of  "A robed figure" and "M16" are missing in the text summaries of compared models. In contrast, our SITA model generates a more accurate summary with the help of pseudo captions containing these essential facts,  which also assists in identifying the salient image correctly.



\section{Rouge-2 of Simple Summaries}

We only plot Rouge-1  and -L scores of simple summaries in Figure~\ref{fig_simple_summary} for better visualization in limited space. The trend of Rouge-2 is similar to that of Rouge-1, as shown in Figure~\ref{fig4}
\end{document}